\documentclass[letterpaper]{article} 
\usepackage{aaai23}  
\usepackage{times}  
\usepackage{helvet}  
\usepackage{courier}  
\usepackage[hyphens]{url}  
\usepackage{graphicx} 
\usepackage{natbib}  
\usepackage{caption} 
\usepackage{algorithm}
\usepackage{algorithmic}
\usepackage{booktabs}

\usepackage{newfloat}
\usepackage{listings}
\DeclareCaptionStyle{ruled}{labelfont=normalfont,labelsep=colon,strut=off} 
\floatstyle{ruled}
\newfloat{listing}{tb}{lst}{}
\floatname{listing}{Listing}
\title{DFEE: Interactive DataFlow Execution and Evaluation Kit}
\author {
    Han He\thanks{Work done during an internship at AWS AI Labs.}\textsuperscript{\rm , 1},
    Song Feng\thanks{Corresponding author.}\textsuperscript{\rm , 2},
    Daniele Bonadiman\textsuperscript{\rm 2},
    Yi Zhang\textsuperscript{\rm 2},
    Saab Mansour\textsuperscript{\rm 2}
}
\affiliations {
    \textsuperscript{\rm 1} Emory University\\
    \textsuperscript{\rm 2} AWS AI Labs \\
    han.he@emory.edu, \{sofeng, dbonadim, yizhngn, saabm\}@amazon.com
}

\usepackage{bibentry}

\begin{document}

\maketitle

\begin{abstract}
DataFlow has been emerging as a new paradigm for building task-oriented chatbots due to its expressive semantic representations of the dialogue tasks. Despite the availability of a large dataset SMCalFlow and a simplified syntax,
the development and evaluation of DataFlow-based chatbots remain challenging due to the system complexity and the lack of downstream toolchains. In this demonstration, we present \textbf{DFEE}, an interactive \textbf{D}ata\textbf{F}low \textbf{E}xecution and \textbf{E}valuation toolkit that supports execution, visualization and benchmarking of semantic parsers given dialogue input and backend database.
 We demonstrate the system via a complex dialog task: event scheduling that involves temporal reasoning. It also supports diagnosing the parsing results via a friendly interface that allows developers to examine dynamic DataFlow and the corresponding execution results. 
To illustrate how to benchmark SoTA models, we propose a novel benchmark that covers more sophisticated event scheduling scenarios and a new metric on task success evaluation. The codes of DFEE have been released on \url{https://github.com/amazon-science/dataflow-evaluation-toolkit}.
\end{abstract}

\section{Introduction}

\begin{figure}[htbp!]
\centering
\includegraphics[width=1\columnwidth]{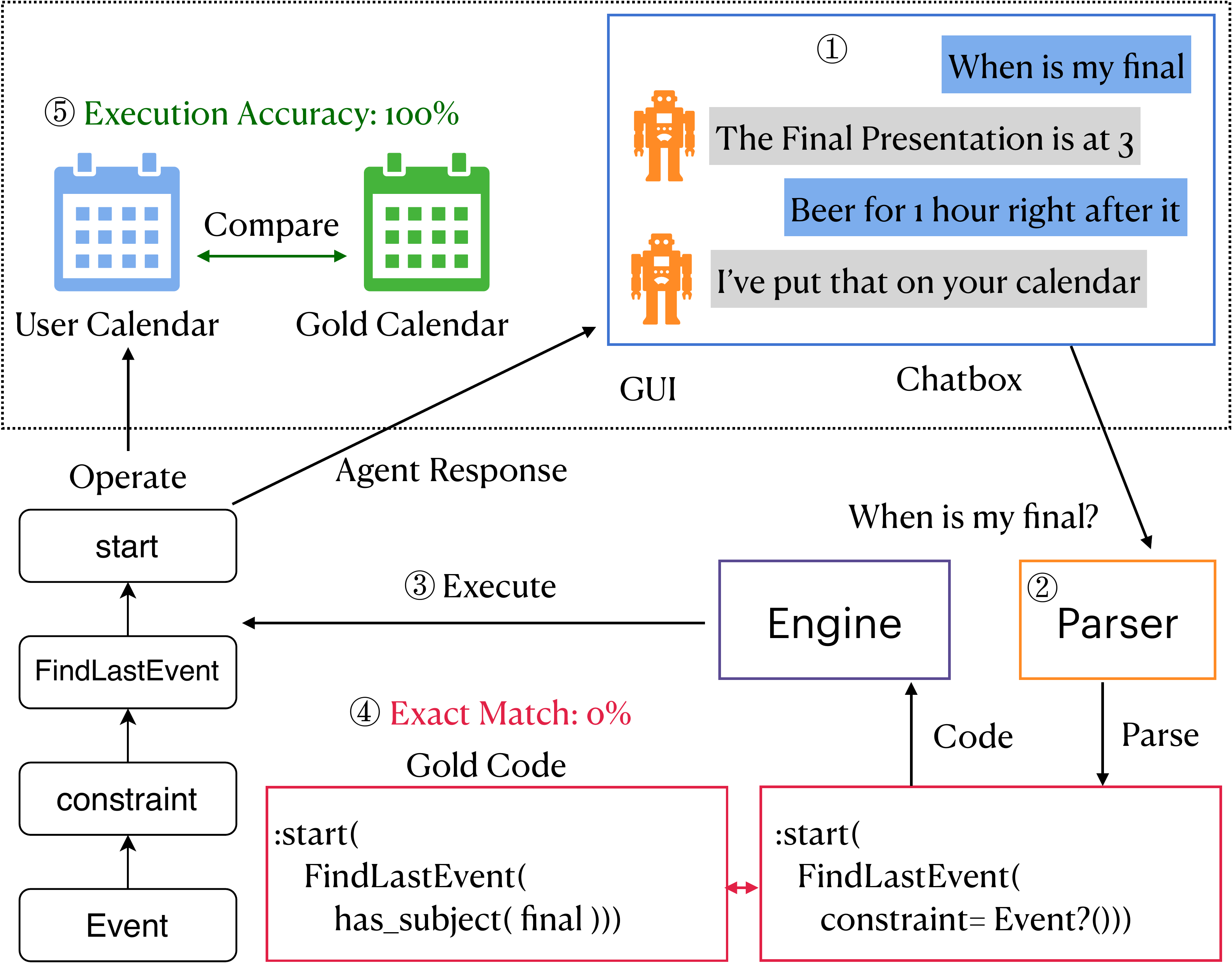}
\caption{DFEE modules: \textcircled{\scriptsize{1}} Chat, \textcircled{\scriptsize{2}} Dialog2API, \textcircled{\scriptsize{3}} API Execution, \textcircled{\scriptsize{4}} Exact Match, \textcircled{\scriptsize{5}} Execution Accuracy.}
\label{ui}
\end{figure}

The representation of dialogue states and agent actions plays a key role in task-oriented dialogue systems as it defines user requests and 
agent behaviors. Popular representations include intent slots \cite{225939}, continuous representations \cite{bordes2017learning}, program synthesis \cite{PGL-010} and Dialog2API \cite{chen-etal-2020-low}.
Fixed symbolic representations are more trainable while distributed continuous representations are more extendable. 
As a structured yet flexible representation, DataFlow composes functions and data into a computational graph 
that interactively gets generated, revised and executed \cite{SMDataflow2020}.
It has been shown to outperform traditional methods on compositional actions, complex anaphora and 
exception handling. 
Though lots of prominent features complemented DataFlow, they inevitably increase the complexity of downstream 
implementations. 
 Recently, \cite[]{meron:2022:ISA18} proposes  OpenDF to simplify SMCalFlow \cite[]{SMDataflow2020} annotations for a more readable syntax. However, developers still cannot directly use OpenDF to generate executable Python programs for a downstream task due to lack of a complete toolchain.

We present the DFEE toolkit, an interactive system that supports execution, evaluation and visualization of DataFlow graphs for event scheduling tasks. Our system allows developers to interact using natural language utterances and to inspect 
the executed DataFlow graphs along with responses in real time. 
We demonstrate our system based on a real-world application of DataFlow-based chatbot, virtual assistant for event scheduling that requires temporal reasoning. To better assess the DataFlow, we enhance the evaluation by supporting a new metric and a novel benchmark dataset that is challenging to SoTA models. 

Our contributions are therefore three-fold: (1) we release a live DataFlow chatbot that allows researchers and practitioners to compile human language into DataFlow graphs and observe realtime responses; (2) we also propose and support execution accuracy as a metric on DataFlow for end-to-end evaluation of semantic parsers; (3) we demonstrate our benchmarking toolkit on a task and a new dataset that challenge the SoTA models.


\section{Evaluation}
Prior work adopts Exact Match (EM) as the sole metric to evaluate DataFlow programs \citep{BenchCLAMP2022}. However, we observe that the same semantics could be represented with different programs due to API implementations or execution processes. For instance, a simple temporal phrase ``10 at night" could be parsed to 3 different APIs from SMCalFlow: \texttt{NumberPM(10)}, \texttt{HourMilitary(22)} or \texttt{HourMinutePm(hour=10, minute=0)}. Thus, we propose a new metric and a dataset to demonstrate the necessity of evaluating end-to-end execution results via our system.

\paragraph{Metrics} To better assess the DataFlow programs, we propose to support a new metric - Execution Accuracy (EA). It is defined as the portion of turns correctly executed. A correctly executed turn has to make the agent generate the same response as the oracle response and ensure the backend database matches the oracle one. 


\paragraph{Benchmark} To support a more complete evaluation of both EM and EA on the exemplar task, a dataset consisting of 1000 dialogues is created by appending temporal reasoning utterances to SMCalFlow dialogues. It provides the annotations on the DataFlow graphs and the expected execution results based on a database as an oracle. Specifically for the dataset in the calendar domain, we design a set of calendar events before and after the execution per each turn. 


\section{Model}

We train and evaluate seq2seq \cite{BenchCLAMP2022} and transition parsers \cite{zhou-etal-2022-online} on the BenchCLAMP \cite{BenchCLAMP2022} split of SMCalFlow. 
Both kinds of parsers take as input the last turn and the current user utterance.
Another BART model (BART-Ours) is also trained on OpenDF simplified version to evaluate the impact of program syntax.

\begin{table}[htbp!]
\centering
{
\begin{tabular}{ccccc}
    \toprule[1pt]
\bf{Model}                               & \bf{Overall} & \bf{Non-Tmp} & \bf{Tmp} & \bf{Sec} \\
\hline
T5\textsuperscript{\textdagger}        & 80.2    & 80.4         & 53.3     & 27.2     \\
BART \textsuperscript{\textdagger}     & 81.0    & 81.2         & \textbf{60.0}     & 1.7      \\
RoBERTa \textsuperscript{\textdaggerdbl} & 76.5    & 76.6         & 53.8     & \textbf{0.1}      \\
BART-Ours               & \textbf{82.9}    & \textbf{83.1}        & 53.3     & 1.1     \\
\bottomrule[1pt]
\end{tabular}
}
\caption{
EM accuracy and latency in seconds (sec) on BenchCLAMP testset, non-temporal (Non-Tmp) and temporal (Tmp) turns. \textdagger\, and \textdaggerdbl\, indicate our experiments of seq2seq  \cite{BenchCLAMP2022} and transition \cite{zhou-etal-2022-online} models respectively. All transformers are large versions.
}
\label{tbl:exact-match}
\end{table}

As shown in Table~\ref{tbl:exact-match}, BART-Ours outperforms BART by 1.9, aligning with the finding by \citet{meron:2022:ISA18}. The transition based parser RoBERTa is optimized for latency but not accuracy. So, we opt for BART-Ours for the later experiments and use it as the default parser in the DFEE system.

\section{System}

We modularize the system into three components: semantic parser, execution engine, and user interface. These components are connected via HTTP requests such that they are swappable. This decoupled design enables the developers to plug-and-play new parser components for benchmarking regardless of the model or programming language.

We build our execution engine upon the OpenDF framework and enhance it in two ways to significantly improve its usability: (1) adding the fulfillment of the necessary APIs for event scheduling tasks; (2) adding worker processes to support multiple concurrent access to the system. 

Our system provides two interfaces: (1) one is the conversational GUI with the visualization of parsing results, i.e., converting dialogue input to an executable program (APIs); (2) one is for CLI on benchmarking our proposed Execution Accuracy of the parsing results.

\paragraph{GUI for Dialog2API} As illustrated in Figure~\ref{ui}, the flow of our GUI is outlined below: one enters an utterance in the chat; the utterance is parsed into a piece of DataFlow code (Dialog2API); the code is executed and the resulting DataFlow graph is visualized; the updates to user's database are reflected to the calendar; the agent response generated during execution is displayed on the chat.

\paragraph{CLI for evaluation} We provide a CLI for benchmarking EM and end-to-end EA of a model. To benchmark a new model, the developers can either provide an online RESTful API URL, or offline parsed DataFlow expressions. The expressions (either parsed or provided) will be executed on our system and the execution and evaluation results will be returned. To demonstrate its functionality, we benchmark BART-Ours on SMCalFlow and our own dataset in Table~\ref{tbl:execution-accuracy}.

\begin{table}[tbp!]
\centering\resizebox{\columnwidth}{!}{
\begin{tabular}{ccccccc}
    \toprule[1pt]
            & \multicolumn{2}{c}{\bf{SM Non-Tmp}} & \multicolumn{2}{c}{\bf{SM Tmp}} & \multicolumn{2}{c}{\bf{Our Tmp}} \\
\bf{Operation}    & EM      & EA     & EM      & EA & EM      & EA   \\
\hline
all         & 80.5 & 84.5 & 66.3 & 56.5 & 3.7 & 28.4  \\
Create & 84.5 & 87.4 & 78.0 & 69.5& 7.6 & 24.7  \\
Query   & 77.8 & 95.9 & 56.5 & 30.4 & 3.3 & 26.4  \\
Update & 45.7 & 71.4 & 25.0 & 25.0 & 2.2 & 43.7  \\
Delete & 75.0 & 86.4 & 0.0  & 50.0 & 0.6 & 12.8 \\
Others      & 71.5 & 68.3 & N/A & N/A  & N/A & N/A   \\
\bottomrule[1pt]    
\end{tabular}
}
\caption{
EM and EA of BART-Ours on SMCalFlow (SM) and our dataset focused on temporal reasoning (Our Tmp).
}
\label{tbl:execution-accuracy}
\end{table}

The model performed significantly worse on the temporal turns, especially on our challenging benchmark that contains diverse scenarios and conversational language styles, appealing for future attention to temporal reasoning and robustness.
EA for each operation is generally higher than EM accuracy, due to the fact that there are many possible ways to express the same semantics, which indicates that EM might be insufficient for evaluating DataFlow programs.
 
 \section{Conclusion}
 
 We presented a novel DataFlow execution and evaluation kit, DFEE. This interactive system has shown to be user-friendly and extendable due to its modularized structure. We further proposed Execution Accuracy metric for SMCalFlow and benchmarked a SoTA parser on a richly annotated dataset. We hope our system can facilitate the future development of robust semantic parsers and intelligent dialogue systems.

\newpage
\bibliography{aaai23}

\end{document}